\documentclass{article}

\usepackage[preprint]{neurips_2026}

\usepackage{times}
\usepackage{soul}
\usepackage{url}
\usepackage[hidelinks]{hyperref}
\usepackage[utf8]{inputenc}
\usepackage[small]{caption}
\usepackage{graphicx}
\usepackage{amsmath}
\usepackage{amsthm}
\usepackage{booktabs}
\usepackage{algorithm}
\usepackage{algorithmic}
\usepackage{listings}
\lstdefinelanguage{json}{
  basicstyle=\ttfamily,
  showstringspaces=false,
  breaklines=true,
  breakatwhitespace=false,
  columns=fullflexible,
  keepspaces=true,
  morestring=[b]",
  stringstyle=\ttfamily,
  literate=
   *{0}{{0}}1 {1}{{1}}1 {2}{{2}}1 {3}{{3}}1 {4}{{4}}1
    {5}{{5}}1 {6}{{6}}1 {7}{{7}}1 {8}{{8}}1 {9}{{9}}1
    {:}{{:}}1 {,}{{,}}1 {\{}{{\{}}1 {\}}{{\}}}1 {[}{{[}}1 {]}{{]}}1
}
\usepackage{xcolor}
\usepackage{wrapfig}
\usepackage{multirow}
\usepackage{enumitem}

\title{Uncertainty Reasoning with Large Language Models for Explainable Disease Diagnosis}

\author{
    Xiaoyang Fan\textsuperscript{1},
    Yufan Cai\textsuperscript{1},
    Zhe Hou\textsuperscript{2} and 
    Jin Song Dong\textsuperscript{1}
    \\
    \textsuperscript{1}National University of Singapore\\
    \textsuperscript{2}Griffith University\\
    \texttt{\{x.fan, cai\_yufan\}@u.nus.edu}\\
    \texttt{z.hou@griffith.edu.au, dcsdjs@nus.edu.sg}
}

\begin{document}

\maketitle
\begin{abstract}
Clinical decision-making often involves reasoning over incomplete, imprecise, and linguistically expressed patient narratives. 
While large language models (LLMs) excel at extracting latent information from natural language, they lack verifiability and interpretability essential for trustworthy medical AI.
We propose a neuro-symbolic reasoning framework that aligns LLMs with formal logic to enable explainable and formally verifiable medical diagnosis. Our approach begins by embedding patient descriptions and clinical guidelines into a neural knowledge base, where LLMs extract structured medical entities, temporal relations, and fuzzy symptom patterns. These are decoded into a symbolic knowledge base, expressed in fuzzy logic and declarative rules.
We then perform two-stage reasoning: (1) inductive symbolic generalization to capture diagnostic patterns from encoded narratives, and (2) inference verification via a logic programming engine to derive and validate diagnoses consistent with clinical standards. Each symptom is treated as a fuzzy predicate with probabilistic weights, and inference paths are auditable, adjustable, and compatible with physician feedback.
Unlike purely statistical methods, our system supports iterative refinement: misalignment between LLM-generated diagnoses and ground-truth can be traced, explained, and corrected through formal rules. 
Besides, our method combines the strengths of logic-based transparency, language models' adaptability, and probabilistic robustness, enabling human-aligned healthcare inference.
The system offers strong generalization with the aid of LLMs and provides verifiable, step-by-step explanations for each reasoning chain.
We validate our framework on public benchmarks, demonstrating its effectiveness in reconciling symbolic reasoning and large language models with real-world clinical narratives.
Our experimental results demonstrate that the proposed framework achieves performance comparable to state-of-the-art LLMs, while additionally offering interpretable reasoning paths and formally verifiable diagnostic conclusions.

\end{abstract}

\section{Introduction}
Clinical diagnosis often requires reasoning under uncertainty: patient narratives may be incomplete, temporally vague, and expressed through qualitative terms, while population-level guidelines must be applied to individual patients whose evidence is partial or conflicting~\cite{han2011uncertainty,bhise2018diagnostic,nasem2015diagnosis}.
Such uncertainty is not merely missing data; it includes ambiguity, complexity, and graded observations such as ``mild fatigue'' or ``high fever'', whose meanings are better represented as degrees than as crisp Boolean facts~\cite{zadeh1996}.
These properties make diagnostic support a setting where accuracy alone is insufficient: the system should expose how symptoms, tests, risks, and exceptions jointly support or weaken each candidate diagnosis.

Large language models (LLMs) have demonstrated strong medical knowledge and language understanding~\cite{singhal2023}, making them useful for extracting latent clinical cues from free-text encounters.
Yet LLM predictions can hallucinate, miscalibrate confidence, and hide the evidence path behind opaque model internals~\cite{ji2023,guo2017calibration}.
This is especially problematic in clinical decision support, where physicians need case-level explanations, traceability, and a way to inspect disagreement between model output and clinical judgment~\cite{amann2020explainability}.
Purely symbolic systems offer transparent rules and verifiable inference, but are brittle when faced with vague symptom descriptions, noisy observations, and incomplete records.
Neuro-symbolic AI provides a natural direction for combining neural language understanding with logical reasoning and explainability~\cite{garcez2023neurosymbolic}, but existing methods often leave fuzzy clinical evidence, probabilistic ranking, and physician-in-the-loop rule revision weakly connected.

To bridge this gap, we propose a \textbf{neuro-symbolic reasoning framework} for interpretable and verifiable clinical diagnosis.
The framework uses LLMs to extract structured entities, temporal cues, and weighted symptom triples from natural-language records; maps these representations into fuzzy predicates and symbolic rules; and performs probabilistic, Prolog-based reasoning that can fuse retrieved case evidence and epidemiological priors to rank diagnostic hypotheses.
This division keeps the language model responsible for perception and normalization, while the symbolic layer retains the clinical commitments that must be checked, revised, and explained.
Instead of treating the rule base as fixed, the system supports a neural-symbolic cycle in which clinicians can edit rules after inspecting the reasoning trace, while recurring case evidence can adjust symptom weights or add and prune rule conditions.
The result is a diagnostic pipeline whose outputs are not only ranked by confidence but also accompanied by auditable inference paths, making uncertainty explicit rather than burying it inside a single generated answer.

This work makes the following key contributions:
\begin{itemize}
    \item We design a hybrid clinical reasoning architecture that connects LLM extraction, fuzzy symptom representation, symbolic deduction, and probabilistic ranking in one pipeline.
    \item We introduce an update mechanism that keeps the symbolic knowledge base inspectable while allowing physician feedback and localized case evidence to revise rule weights and structure.
    \item We evaluate the framework on public clinical benchmarks, showing that the hybrid design improves robustness and explainability over symbolic ablations and remains competitive with strong LLM baselines.
\end{itemize}

By grounding neural extraction in explicit fuzzy and symbolic representations, our framework offers a practical route toward diagnostic AI that can reason over ambiguous patient narratives while preserving traceability for clinical review.
\section{Motivating Example}
\label{subsec:motivating_example}
This section shows a real-world example for our proposed framework.
\paragraph{Free-text clinical note.}
\begin{quote}
\small
``A 58-year-old man reports \textit{on-and-off chest heaviness} for the past ten days, usually after climbing two flights of stairs.  
He describes the discomfort as a ``tight belt'' around the chest, lasting 3–5 minutes and easing with rest.  
He also feels \textit{mild breathlessness} on exertion but denies nausea or diaphoresis.  
Blood pressure at triage is 155/92~mmHg; heart rate 88~bpm.  
ECG shows non-specific ST changes; high-sensitivity troponin is normal.  
The patient has a 30-pack-year smoking history, treated hypertension, and both parents suffered heart attacks in their 50s.''%
\end{quote}

The note poses several challenges: \textbf{Temporal vagueness}\textemdash{}phrases such as “past ten days” and “3--5~minutes” are qualitative spans that symbolic systems find hard to encode; \textbf{Graded descriptors}\textemdash{}words like \emph{on\,-and\,-off}, \emph{tight belt}, and \emph{mild} convey fuzzy severities rather than crisp categories; \textbf{Hidden hypotheses}\textemdash{}no explicit mention of ``stable angina'' or ``coronary artery disease'' appears and the link must be inferred; \textbf{Mixed evidence}\textemdash{}normal troponin and nonspecific ECG changes argue against acute infarction yet do not rule out ischaemia, demanding probabilistic reasoning; and \textbf{Clinical accountability}\textemdash{}cardiologists require a transparent chain that explains \emph{why} the system recommends further work-up or discharge.

Our framework will adopt the following procedure to process this case:
\begin{enumerate}[leftmargin=*]
  \item \textbf{LLM extraction.}
        The language model converts the note into structured snippets such as  
        \texttt{symptom(chest\_pain, heaviness, exertional)},  
        \texttt{duration(chest\_pain, 10\,days)},  
        \texttt{risk(smoking)}, \texttt{risk(family\_history)}, \texttt{test(ecg\_nonspecific)}, \texttt{lab(troponin\_normal)}.
  \item \textbf{LLM-as-a-Judge verification.}  
        A secondary pass checks each snippet against the original text to remove hallucinations and resolve contradictions.  
        For example, the judge confirms that \texttt{symptom(chest\_pain, exertional)} is valid, but would reject a hallucinated item like \texttt{symptom(nausea)} as it was denied.
  \item \textbf{Fuzzy symbolisation.}  
        Each snippet becomes a fuzzy fact with a weight, e.g.\ \texttt{fuzzy\_symptom(chest\_pain, 0.8)}. 
  \item \textbf{Rule compilation.}  
        A library of plain-text guideline knowledge includes rules such as:
\begin{flushleft}
\small\ttfamily
if chest\_pain $\wedge$ exertional $\wedge$ risk\_factors $\to$ investigate\_for\_angina\\
if troponin\_normal $\wedge$ ecg\_nonspecific $\to$ low\_prob\_acute\_MI
\end{flushleft}
  \item \textbf{Weighted reasoning.}  
        The system combines fuzzy facts with rules to score differential diagnoses.  
        In this case it outputs:  
        \texttt{stable\_angina (0.72)}, \texttt{noncardiac\_chest\_pain (0.18)}, \texttt{acute\_MI (0.10)}.
  \item \textbf{Prolog-based formal inference}  
        The same facts and rules are exported into a Prolog program for rigorous inference:
\begin{lstlisting}[language=prolog,basicstyle=\ttfamily\small]
% --- facts (imported) ---
symptom(chest_pain).
trigger(exertion).
...
risk(smoking).
...
lab(troponin_normal).

% --- rules (subset) ---
diagnosis(stable_angina) :-
    symptom(chest_pain),
    trigger(exertion),
    risk(_),       
    \+ lab(troponin_elevated).
diagnosis(acute_mi) :-
    symptom(chest_pain),
    lab(troponin_elevated).

?- diagnosis(D).    % query returns D = stable_angina.
\end{lstlisting}

  \item \textbf{Explanation trace.}  
        A human-readable summary is generated:

\begin{quote}
\small
\textbf{Why stable angina?} Chest pain is exertional and heavy; patient has hypertension and smoker; family history present. Normal troponin lowers acute MI probability but does not negate stable ischaemia. Guideline rule XX triggered.
\end{quote}

  \item \textbf{Physician feedback loop.}  
        If the cardiologist remarks that the pain quality seems atypical, a single click down-weights \texttt{chest\_pain, heaviness}.  
        The system re-runs logic instantly, updating the angina score to 0.55 and recommending a treadmill test instead of immediate angiography.
  \item \textbf{Automated rule update.}
        If pleuritic chest pain is observed repeatedly with pulmonary embolism and rarely with stable angina, the system automatically adds \texttt{pleuritic\_pain} to the embolism rule set and removes it from angina.
        If exertional heaviness with smoking and family history is consistently present in true angina cases, the system retains and highlights this combination as a key trigger.

\end{enumerate}

\vspace{-0.5em}

This example shows that our approach \emph{(i)} keeps flexibility for messy language, \emph{(ii)} converts it into weighted symbolic facts, \emph{(iii)} applies clear text rules for inference, and \emph{(iv)} exposes every step for clinician audit—capabilities absent from purely symbolic or purely neural systems.
\begin{figure}[ht]
    \centering
    \includegraphics[width=0.8\linewidth]{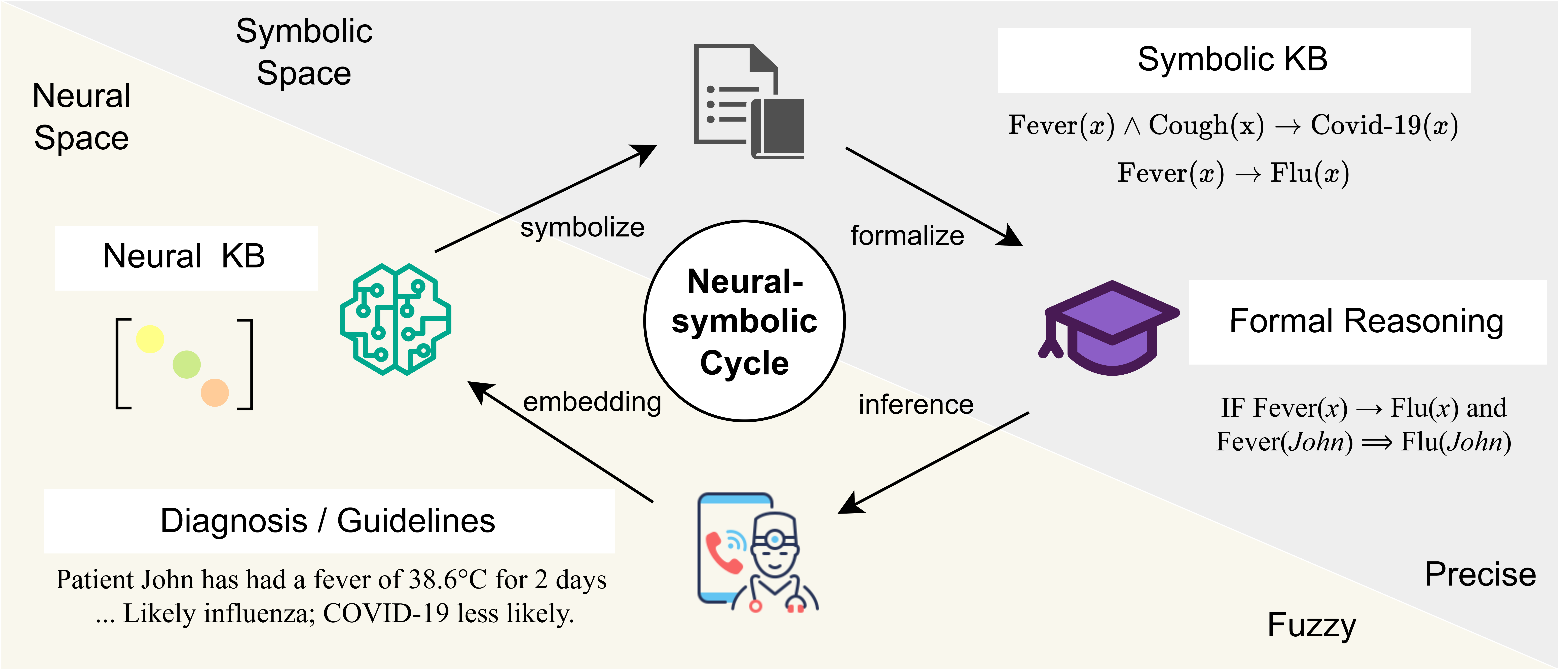}
    \caption{The Neuro-symbolic Cycle.}
    \label{fig:KG}
\end{figure}

\section{Approach}
\begin{figure}[ht]
    \centering
    \includegraphics[width=0.9\linewidth]{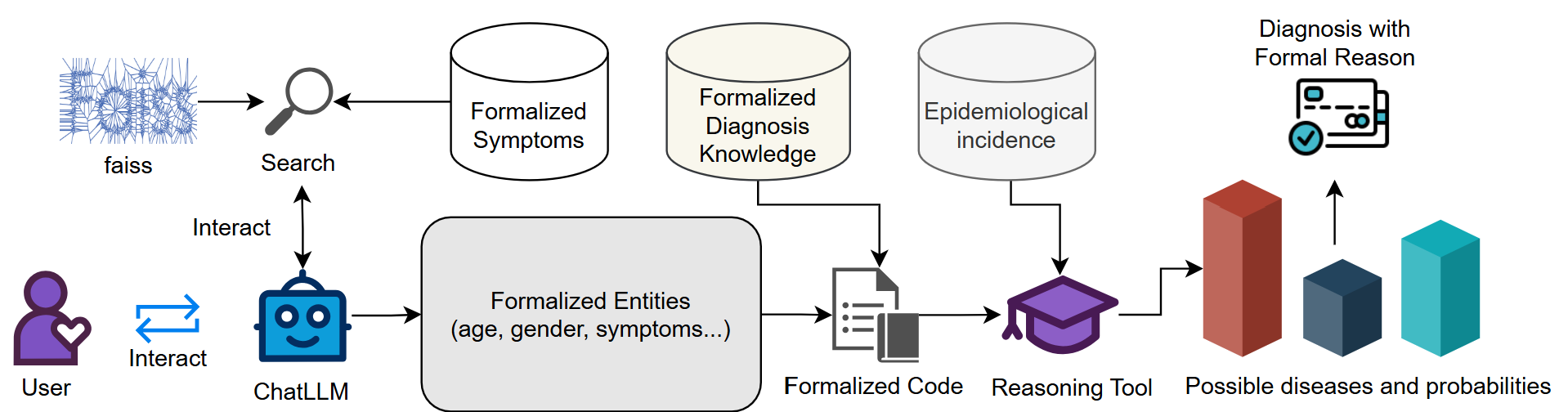}
    \caption{The Medical Diagnosis Framework}
    \label{fig:framework}
\end{figure}

Our framework comprises three tightly coupled modules:  
\textbf{(A)}~formal knowledge construction and reasoning toolchain,
\textbf{(B)}~a neuro-symbolic learning cycle that evolves the rule base through both data-driven updates and physician feedback, as shown in ~\autoref{fig:KG}, and  
\textbf{(C)}~a real-time diagnostic engine.  
~\autoref{fig:framework} illustrates the complete workflow.

\subsection{Formal Knowledge Construction}
\label{subsec:knowledge}

\paragraph{Neural embedding of clinical corpora.}
We aggregate free-text sources—clinical notes, guideline paragraphs, and textbook explanations—and encode them with a domain-tuned LLM, yielding dense vectors that preserve semantic relations across symptoms, findings, and diseases.

The symbolic rule set \(\mathcal{R}\) is \emph{not} fixed: we induce and
refine rules directly from corpora using a prompting strategy.
Formally, for each sentence–diagnosis pair \((\mathbf{x},D^{\star})\) the model \(\mathcal{M}_{\theta}\) is prompted to propose Horn templates
\(T = (H \leftarrow B_{1},\dots,B_{n})\),
where \(H\) is a diagnosis literal and \(B_{i}\) are symptom or lab predicates.
A template score
\(\text{Score}(T)=\sigma\!\bigl(\mathcal{M}_{\theta}(\mathbf{x},T)\bigr)\)
is averaged over a minibatch and compared with a threshold \(\tau\);
templates with \(\text{Score}(T)>\tau\) are concretised and added to the rule set \(\mathcal{R}\),
which thus evolves on-the-fly from new data and clinician feedback.

\paragraph{Symbolic decoding via LLM prompting.}
A constrained decoding prompt asks the same LLM to transform each embedded segment into a structured triple set  
\texttt{\{entity, relation, value\}}; for example  
\texttt{(chest\_pain, severity, intermittent)} or  
\texttt{(troponin, level, normal)}.  
The triples are mapped to \emph{fuzzy predicates} whose truth degrees $w\in[0,1]$ reflect linguistic hedges (e.g.\ ``mild'', ``occasional'').

\paragraph{Prolog rule synthesis.}
We transform triples into declarative rules using a small template library:
\begin{flushleft}\ttfamily\small
symptom(\$X) :- text\_mention(\$X), weight(\$W), ...\\
diagnosis(...) :- symptom(...), trigger(...), risk\_factor(...).\\
\end{flushleft}
Resulting rules and fuzzy facts together constitute a \emph{symbolic knowledge base}.

\paragraph{Symbolic Trace Generation.}
\label{subsec:trace}

For each inferred hypothesis \(H\), we generate a
\emph{fine‐grained proof tree} \(\mathcal{T}_H\) with LLM:
the prompt includes the nodes, which are instantiated predicates,
and the edges, which are rule applications, and the leaf weights, which are the
fuzzy activations \(\tilde{B}_i\).
The overall confidence is
\[
\mathrm{Conf}(H) \;=\;
\sum_{p \in \mathcal{P}_H} \Bigl[
    \prod_{i\in p} \tilde{B}_i
\Bigr],
\]
where \(\mathcal{P}_H\) enumerates all root‐to‐leaf paths in \(\mathcal{T}_H\).

\subsection{Neural-Symbolic Cycle}
\label{subsec:rule_update}

\paragraph{Physician‐Driven Rule Editing.}
\label{subsec:rule_edit}

A domain expert may inject a new rule
\(\rho^\mathrm{new}\) with weight \(w^\mathrm{new}\).
The system recomputes
\(\mathcal{R} \leftarrow \mathcal{R} \cup \{\rho^\mathrm{new}\}\),
re-verifies consistency,
and stores \(\mathcal{K}^{(t+1)}\).

After any diagnostic session, discrepancies between the system’s conclusion and ground-truth labels are surfaced in a visual diff tool. 
Clinicians can (i) add an exception rule, (ii) adjust a predicate weight, or (iii) supply counter-examples.  
Edits are versioned; the KB is re-compiled, closing an iterative \emph{neuro-symbolic learning loop}.

\paragraph{Automated Rule Updating.}
The symbolic rule base \(\mathcal{R}\) can also be continuously refined from case data.  
We employ a two–stage mechanism: (i) an online passive–aggressive ranking update that adjusts weights when ground–truth diseases are mis–ranked, and (ii) structural add/prune operations guided by evidence counts.

\paragraph{Online PA ranking update.}
Given a case with observed symptoms \(x=\{s_1,\dots,s_n\}\) and ground–truth diseases \(G\subseteq\mathcal{D}\),
each disease candidate \(d\) receives a score
\[
\text{score}(d\mid x) = \sum_{s\in x} w_{d,s},
\]
with \(w_{d,s}\in(0,1]\).
If a true disease \(d^+\in G\) scores below a false disease \(d^-\in\hat{\mathcal{D}}_K\setminus G\),
the margin loss is
\[
\ell = \max\Bigl(0,\; m - \bigl(\text{score}(d^+\mid x) - \text{score}(d^-\mid x)\bigr)\Bigr),
\]
where \(m\) is the desired ranking margin.
The update step is
\[
\tau = \min\!\left(C,\; \frac{\ell}{2|x|}\right),
\]
where \(C\) is an aggressiveness cap controlling the maximum per-case update, and \(|x|\) is the number of active symptoms in the case.
For each symptom \(s\in x\):
\[
w_{d^+,s} \leftarrow \text{clip}_{[0,1]}(w_{d^+,s}+\tau), \qquad
w_{d^-,s} \leftarrow \text{clip}_{[0,1]}(w_{d^-,s}-\tau).
\]
A zero-to-positive update effectively \emph{adds} a symptom to a disease; values clipped to zero are flagged for deletion.

\paragraph{Structure thresholds.}
Beyond weight nudges on individual cases, we periodically inspect symptom-disease edges against aggregated statistics to decide whether to keep, add, or prune them.
We maintain counts \(c^+_{d,s}\) (cases with disease \(d\) and symptom \(s\)) and
\(c^-_{d,s}\) (cases without \(d\) but with \(s\)).
Define the smoothed specificity ratio
\[
r_{d,s}=\frac{c^+_{d,s}+1}{c^-_{d,s}+1}.
\]
Edges are updated by simple rules:
\[
\text{Add }(d,s)\;\;\text{if}\;\;c^+_{d,s}\ge m_{\text{pos}}\;\;\wedge\;\;r_{d,s}\ge\rho_a,
\]
\[
\text{Prune }(d,s)\;\;\text{if}\;\;w_{d,s}=0\;\;\wedge\;\;r_{d,s}<\rho_p,
\]
where \(m_{\text{pos}}\) is minimum number of positive co-occurrences required and \(\rho_a\) and \(\rho_p\) are the specificity thresholds for adding and pruning respectively.

\paragraph{Integration.}
This procedure ensures that \(\mathcal{R}\) evolves automatically with new data:
weights adapt case by case, while the structure of supporting symptoms
is expanded or simplified.  All updates are versioned alongside
clinician edits, producing a transparent
audit trail of rule evolution.

\paragraph{Versioned Knowledge Base.}
\label{subsec:versioning}

Every update to \(\mathcal{R}\) or the fuzzy membership set
\(\{\mu_k\}\) produces a new \emph{snapshot}
\(\mathcal{K}^{(t)} = \langle \mathcal{R}^{(t)}, \{\mu_k^{(t)}\}\rangle\),
time‐stamped at \(t\).
During inference we deploy the latest \(\mathcal{K}^{(t)}\),
but we retain historical snapshots to enable
\emph{counterfactual audits}:
clinicians may query why a diagnosis changed between
\(\mathcal{K}^{(t-1)}\) and \(\mathcal{K}^{(t)}\).
The diff tool highlights rule additions, deletions, and weight shifts,
providing a transparent evolution trail.

\subsection{Real-time Diagnostic Engine}
\label{subsec:inference}

This module turns a free-text encounter note into a probabilistically ranked differential diagnosis.

\paragraph{Note segmentation \& temporal alignment.}
Before any neural encoding, the note is split into \textit{clinical
segments} (chief complaint, history, vitals, labs) with the interaction of LLM.  
Time expressions such as
``for the past week’’ or ``resolved yesterday’’ are normalised to
relative timestamps; symptoms are tagged with the most recent timestamp
to allow time-aware rule checks (e.g.\ ``acute’’ vs ``chronic’’).

\paragraph{Case vectorisation and retrieval.}
Each segment is embedded by the same domain LLM encoder
used in Module A.  The full note vector \(\mathbf{v}\) is obtained by
mean pooling.  We query a FAISS index of \(\sim\)100 k de-identified
cases and retrieve the \(k\) nearest neighbours
\(\mathcal{N}_k\).
If the neighbours span multiple diagnostic clusters, \(k\) is
adaptively reduced until the Gini impurity of diagnoses in
\(\mathcal{N}_k\) falls below 0.3, yielding a more coherent context
set for weighting.

\paragraph{Hybrid symptom weighting.}
Let \(s_i\) be a unique symptom concept.
\begin{itemize}[leftmargin=1.8em,itemsep=1pt]
\item \textbf{Neighbour prior.}
   Each \(s_i\) inherits a prior
   \(w_i^{\text{retr}} = \max_{c\in\mathcal{N}_k}
       \bigl(\cos(\mathbf{v},\mathbf{v}_c)\bigr)\),
   reflecting the most similar neighbour expressing that symptom.
\item \textbf{Intrinsic salience.}
   The LLM parser attaches
   \(w_i^{\text{text}}\in[0,1]\) from hedges (\emph{mild}, \emph{severe}),
   frequency cues, and section headers.
\item \textbf{Calibration.}
   We fuse both sources via a softmax blend
   \[
     w_i = \frac{\exp(\alpha\,w_i^{\text{text}}) +
               \exp(\beta\,w_i^{\text{retr}})}%
              {\sum_j
               \bigl[\exp(\alpha\,w_j^{\text{text}})+
                     \exp(\beta\,w_j^{\text{retr}})\bigr]},
   \]
   with \(\alpha{=}\beta{=}3\) by default; clinicians may
   shift the slider to privilege text or retrieval evidence.
\end{itemize}

\paragraph{Fuzzy Prolog reasoning.}
The weighted facts
\texttt{fuzzy\_symptom($s_i$, $w_i$)} are asserted in SWI-Prolog
augmented with the \texttt{library(fuzzy)} package.
Rules compiled in Module A fire if their t-norm activation exceeds
\(\gamma=0.4\).
The solver outputs a candidate list
\(\mathcal{D}=\{(d_j,\rho_j)\}\)
where \(\rho_j\) is the maximum rule activation supporting \(d_j\).

\paragraph{Epidemiological prior fusion.}
We retrieve prevalence \(\pi_j\) from the knowledge base,
conditioned on \{\textit{age, sex, region}\}.
Posterior probabilities follow
\[
  P(d_j\mid\mathbf{s}) =
    \frac{\rho_j\,\pi_j}{\sum_{d'} \rho_{d'}\,\pi_{d'}}.
\]
Rare, high-risk conditions (e.g.\ aortic dissection) retain
visibility because \(\rho_j\) can outweigh low \(\pi_j\).

By unifying neural extraction with symbolic inference, our approach maintains the adaptability of LLMs while guaranteeing formally verifiable reasoning chains, fulfilling the dual requirements of accuracy and trustworthiness in clinical AI.

\section{Experiments}
We conducted a series of experiments to evaluate the effectiveness of our neuro-symbolic reasoning framework on clinical inference tasks.
Our evaluation focused on the correctness of the diagnosis output.
To guide our experiments, we addressed the following research questions:
\begin{itemize}
\item \textbf{RQ1:} How accurate is our hybrid reasoning system in performing diagnostic inference compared to baselines?
\item \textbf{RQ2:} Does the incorporation of fuzzy quantification and probabilistic inference improve diagnostic accuracy?
\item \textbf{RQ3:} How consistent are the intermediate symptoms in terms of error rates and how consistent are the final reasoning explanations?
\item \textbf{RQ4:} How does our system compare to baselines in terms of token usage, cost, and runtime efficiency?
\end{itemize}

\subsection{Setup}
We evaluated our system on two publicly available benchmark datasets from Hugging Face~\cite{huggingface} and a health record dataset from PhysioNet~\cite{goldberger2000}:
\begin{itemize}
\item  \textit{gretelai/symptom\_to\_diagnosis}~\cite{gretelai}, containing procedurally generated patient profiles with known ground-truth diagnoses based on rule structures.
This is a single-label dataset, which means that each symptom description corresponds to a single disease.
\item  \textit{lavita/ChatDoctor-iCliniq}~\cite{lavita}, containing doctor-patient dialogues from the public medical QA platform iCliniq, preprocessed into symptom triples or fuzzy descriptors.
This is a multi-label dataset, meaning that each symptom description may correspond to multiple diseases.
\item
\textit{MIMIC-IV} (v3.1)~\cite{johnson2023, johnson2024}, containing structured lab test results and clinical notes written by the doctors.
This is also a multi-label dataset.
\end{itemize}

We compared our system with GPT-4o, o4-mini and DeepSeek-R1 as the baselines.
Additionally, we established three ablation baselines including:
\begin{enumerate}
\item \textit{Symbolic-only}: deduction via crisp rule matching;
\item \textit{Symbolic + Probabilistic}: diagnosis without fuzzy quantification;
\item \textit{Symbolic + Fuzzy}: diagnosis without probabilistic inference.
\end{enumerate}
We also established a simple baseline model that utilizes only the first appearing symptom from the input to predict outcomes, helping to evaluate performance relative to the task.
Evaluation metrics include: Top-1, Top-3 \& Top-5 Accuracy / Precision / Recall / F1-score on known diagnostic labels.

\subsection{Results and Analysis}

~\autoref{tab:merged_s2d_results} presents the Top-k results on the first dataset.
Since it is single-label (each symptom description corresponds to only one true diagnosis), accuracy and recall are identical across all Top-k results, and all four evaluation metrics are identical in the Top-1 results.

\begin{table*}[htp]
\centering
\small
\caption{Top-k results on \textit{symptom\_to\_diagnosis} dataset (all values in \%)}
\begin{tabular}{lccccccccc}
\toprule
& \multicolumn{3}{c}{Top-1} & \multicolumn{3}{c}{Top-3} & \multicolumn{3}{c}{Top-5} \\
\cmidrule(lr){2-4} \cmidrule(lr){5-7} \cmidrule(lr){8-10}
Method & Acc & Prec & F1 & Acc & Prec & F1 & Acc & Prec & F1 \\
\midrule
Simple & 21.8 & - & - & 37.3 & 12.8 & 19.1 & 43.6 & 9.2 & 15.2\\
DeepSeek-R1 & 34.4 & - & - & 52.8 & 17.6 & 26.4 & 59.8 & 12.0 & 20.0 \\
o4-mini & 36.5 & - & - & 52.7 & 17.6 & 26.3 & 63.7 & 12.7 & 21.2 \\
GPT-4o & 39.0 & - & - & 56.7 & 18.9 & 28.3 & 66.6 & 13.3 & 22.2 \\
Sym-only & 37.4 & - & - & 56.1 & 18.7 & 28.1 & 62.9 & 12.6 & 21.0 \\
Sym + Prob & 42.1 & - & - & 61.4 & 20.5 & 30.8 & 69.3 & 13.9 & 23.1 \\
Sym + Fuzzy & 39.8 & - & - & 62.9 & 21.0 & 31.5 & 70.2 & 14.1 & 23.4 \\
Full Hybrid & \textbf{44.8} & - & - & \textbf{65.9} & \textbf{22.0} & \textbf{33.0} & \textbf{73.5} & \textbf{14.7} & \textbf{24.5} \\
\bottomrule
\end{tabular}
\label{tab:merged_s2d_results}
\end{table*}

As shown, our full hybrid method outperformed baselines in all Top-k evaluations (RQ1).
The ablation experiments demonstrated that both the fuzzy quantification module and the probabilistic inference module independently improve the accuracy of the symbolic-only approach, and the hybrid method achieved the highest accuracy (RQ2).

\begin{table}[htp]
\centering
\small
\caption{Top-k results on \textit{iCliniq} dataset (all values in \%)}
\begin{tabular}{lcccccccccccc}
\toprule
& \multicolumn{4}{c}{Top-1} & \multicolumn{4}{c}{Top-3} & \multicolumn{4}{c}{Top-5} \\
\cmidrule(lr){2-5} \cmidrule(lr){6-9} \cmidrule(lr){10-13}
Method & Acc & Prec & Rec & F1 & Acc & Prec & Rec & F1 & Acc & Prec & Rec & F1 \\
\midrule
GPT-4o & 48.6 & - & 34.0 & 40.0 & 68.8 & 26.6 & 55.9 & 36.0 & 79.0 & 19.0 & 66.5 & 29.6 \\
Sym-only & 32.6 & - & 22.8 & 26.9 & 42.6 & 15.3 & 32.1 & 20.7 & 58.2 & 12.6 & 44.4 & 19.7 \\
Sym + Prob & 36.2 & - & 25.4 & 29.8 & 57.8 & 20.3 & 42.7 & 27.5 & 65.2 & 14.2 & 49.9 & 22.1 \\
Sym + Fuzzy & 40.6 & - & 28.4 & 33.4 & 61.0 & 21.3 & 44.7 & 28.8 & 67.4 & 14.8 & 51.7 & 23.0 \\
Full Hybrid & \textbf{42.0} & - & \textbf{29.4} & \textbf{34.6} & \textbf{63.2} & \textbf{22.7} & \textbf{47.6} & \textbf{30.7} & \textbf{71.8} & \textbf{16.0} & \textbf{56.0} & \textbf{24.9} \\
\bottomrule
\end{tabular}
\label{tab:merged_icliniq_results}
\end{table}

~\autoref{tab:merged_icliniq_results} presents the Top-k results on the iCliniq dataset.
Since this dataset is multi-label, only accuracy and precision are identical in the Top-1 results; the other metrics differ in all other cases.
The ablation experiments confirmed that both fuzzy quantification and probabilistic inference improve diagnostic accuracy, consistent with our earlier conclusions (RQ2).

\paragraph{Error Analysis}
As shown, our full hybrid method slightly underperformed GPT-4o in the Top-k evaluations on the iCliniq dataset.
Upon analysing the dataset, we identified the following reasons for the lower accuracy:
\begin{itemize}
\item Some data points' true labels (diseases) were not included in our knowledge base.
\item Some data points lacked explicit diagnostic conclusions from doctors, making their true labels unreliable.
\item Some data points contained patient-provided text that was not symptom descriptions.
\item Some data points included patients who already knew their diagnoses, mentioning disease names directly.
\item Some data points contained non-text inputs, which our model could not process.
\end{itemize}

To mitigate these errors, we excluded these data points, resulting in a trimmed iCliniq dataset.
Since ~\autoref{tab:merged_icliniq_results} already validated the ablation experiments, we directly compared GPT-4o with our full hybrid method on the trimmed dataset without further ablation studies.

\begin{table*}[htp]
\centering
\small
\caption{Top-k results on trimmed \textit{iCliniq} dataset (all values in \%)}
\begin{tabular}{lccccccccccc}
\toprule
& \multicolumn{3}{c}{Top-1} & \multicolumn{4}{c}{Top-3} & \multicolumn{4}{c}{Top-5} \\
\cmidrule(lr){2-4} \cmidrule(lr){5-8} \cmidrule(lr){9-12}
Method & Acc/Prec & Rec & F1 & Acc & Prec & Rec & F1 & Acc & Prec & Rec & F1 \\
\midrule
Simple & 29.3 & 20.3 & 24.0 & 40.0 & 16.2 & 29.8 & 21.0 & 45.0 & 11.9 & 33.8 & 17.6 \\
DeepSeek-R1 & 42.1 & 29.2 & 34.4 & 63.1 & 23.6 & 49.0 & 31.8 & 72.0 & 17.0 & 58.8 & 26.3 \\
o4-mini & 45.4 & 31.5 & 37.2 & 69.6 & 26.1 & 54.3 & 35.3 & 76.5 & 17.9 & 62.2 & 27.9 \\
GPT-4o & 49.9 & 34.6 & 40.8 & 69.8 & 27.2 & 56.6 & 36.8 & 80.3 & 19.5 & 67.4 & 30.2 \\
Full Hybrid & \textbf{47.0} & \textbf{32.6} & \textbf{38.5} & \textbf{70.7} & \textbf{25.4} & \textbf{52.7} & \textbf{34.3} & \textbf{80.3} & \textbf{17.9} & \textbf{62.1} & \textbf{27.8} \\
\bottomrule
\end{tabular}
\label{tab:trimmed}
\end{table*}

\begin{table*}[htp]
\centering
\small
\caption{Top-k results on \textit{MIMIC-IV} dataset (all values in \%)}
\begin{tabular}{lccccccccccc}
\toprule
& \multicolumn{3}{c}{Top-1} & \multicolumn{4}{c}{Top-3} & \multicolumn{4}{c}{Top-5} \\
\cmidrule(lr){2-4} \cmidrule(lr){5-8} \cmidrule(lr){9-12}
Method & Acc/Prec & Rec & F1 & Acc & Prec & Rec & F1 & Acc & Prec & Rec & F1 \\
\midrule
Simple & 32.7 & 20.5 & 25.2 & 39.5 & 18.0 & 28.6 & 22.1 & 43.5 & 13.1 & 31.6 & 18.5 \\
DeepSeek-R1 & 50.1 & 38.8 & 43.7 & 71.8 & 34.3 & 59.9 & 43.6 & 85.6 & 21.5 & 72.1 & 33.1 \\
o4-mini & 51.6 & 40.5 & 45.4 & 74.0 & 35.6 & 62.4 & 45.3 & 86.4 & 23.3 & 74.0 & 35.4 \\
GPT-4o & 53.3 & 42.0 & 47.0 & 75.3 & 36.3 & 63.5 & 46.2 & 88.4 & 24.7 & 76.2 & 37.3 \\
Full Hybrid & \textbf{57.2} & \textbf{45.6} & \textbf{50.7} & \textbf{76.3} & \textbf{38.2} & \textbf{64.4} & \textbf{48.0} & \textbf{89.3} & \textbf{25.0} & \textbf{75.5} & \textbf{37.6} \\
\bottomrule
\end{tabular}
\label{tab:mimic-iv}
\end{table*}

As shown in ~\autoref{tab:trimmed}, on the trimmed dataset, GPT-4o still slightly outperformed our framework.
This aligns with our expectation: In highly unstructured or ambiguous contexts, the native performance of LLM is difficult to surpass.
When symptom descriptions are highly ambiguous, our framework has limited room to outperform GPT-4o because it relies on GPT-4o for symptom extraction.
On the other hand, as shown in ~\autoref{tab:mimic-iv}, our system outperformed GPT-4o and other LLMs on MIMIC-IV. Compared to iCliniq, this dataset has clearer symptom signals, aligning with the strengths of our method.

\paragraph{Consistency Check}
We used GPT-4o to evaluate the error rate of the symptoms (which is the input of our formal reasoning) and the explainability score (0-4) of our output.
The score measures how much our reasoning is consistent and aligns with the case.

\begin{table}[htp]
\centering
\small
\caption{Error rate and explainability evaluation results}
\begin{tabular}{ccc}
\toprule
Dataset & Error Rate of Input & Explainability Score of Output \\
\midrule
\textit{S2D} & 1.28\% & 3.67/4.00 \\
\textit{iCliniq} & 0.88\% & 3.20/4.00 \\
\textit{MIMIC-IV} & 0.65\% & 3.58/4.00 \\
\bottomrule
\end{tabular}
\label{tab:error_explainability}
\end{table}

As shown in ~\autoref{tab:error_explainability}, our system demonstrated high consistency across all evaluated datasets. The error rates of symptom inputs were notably low, indicating reliable input data for our formal reasoning process.
Furthermore, the explainability scores highlighted the robustness of our system’s reasoning explanations.
These results suggest that our framework effectively maintained consistency in both intermediate symptom predictions and final reasoning explanations (RQ3).

\paragraph{Cost Analysis}
In addition to diagnostic accuracy, we evaluated the time and monetary cost of our hybrid reasoning framework compared to GPT-4o on processing 1,000 samples.

\begin{table}[H]
\centering
\small
\caption{Comparison of token usage, monetary cost, and runtime}
\begin{tabular}{lccc}
\toprule
Method & Tokens Used (M) & Cost (USD) & Runtime (min) \\
\midrule
GPT-4o & 5.75 & \$14.68 & 26.61 \\
Full Hybrid (Ours)   & 2.87 & \$7.61  & 48.69 \\
\bottomrule
\end{tabular}
\label{tab:time_cost}
\end{table}

Although our method takes more time, it is significantly more cost-efficient and uses fewer tokens than GPT-4o (~\autoref{tab:time_cost}). 
This highlights the practical viability of our neuro-symbolic system in resource-constrained settings (RQ4).
\section{Discussion}
Our proposed neuro-symbolic reasoning framework demonstrates that clinical diagnosis can benefit from the structured transparency of symbolic logic and the flexibility of LLMs.
By systematically combining these paradigms, our system not only offers interpretable and traceable reasoning chains but also adapts to real-world data.

\paragraph{Limitations}
Despite its strengths, our implementation still has three principal limitations. 
First, the symbolic reasoning layer rests on manually curated rules; while this ensures interpretability, it inevitably overlooks the full diversity of clinical presentations and constrains scalability to new domains or emerging diseases. 
Second, the fuzzy-logic module relies on expert-crafted membership functions derived from fixed clinical thresholds, demanding substantial domain expertise and risking sub-optimal generalization across heterogeneous patient populations unless further data-driven calibration is performed. 
Finally, the framework currently models symptoms as static observations and therefore lacks an explicit temporal representation of their onset and evolution—an essential capability for many real-world diagnostic tasks.


\section{Related Work}
\paragraph{Healthcare AI Systems}
Large language models (LLMs) have been increasingly applied to medical question answering (QA), dialogue systems, and preliminary diagnosis due to their ability to interpret free-text symptom descriptions and extract clinical insights~\cite{lee2023, li2024, singhal2023}. 
Systems such as ChatDoctor~\cite{li2023}, BioGPT~\cite{luo2022}, and GatorTron~\cite{yang2022} support patient–LLM interaction, biomedical knowledge generation, and document-level inference across a range of tasks.
Domain-specific models like Med-PaLM 2~\cite{singhal2025}, trained on curated medical corpora and instruction-tuned with expert feedback, achieve expert-level performance on benchmarks such as MedQA, PubMedQA~\cite{jin2019}, and USMLE-style exams~\cite{jin2021}.
Others, like PMC-LLaMA~\cite{wu2023}, continue this trend by leveraging large-scale biomedical publications to enhance generation quality in medical settings.

These models have demonstrated potential in clinical summarization~\cite{huang2025}, differential diagnosis~\cite{wang2024}, and multi-turn medical dialogue~\cite{kim2024}.
However, they remain prone to factual hallucinations~\cite{ji2023}, spurious associations, and incorrect clinical reasoning~\cite{busch2025}.
LLMs typically lack robust temporal and causal inference capabilities~\cite{sun2013}, and often exhibit poor uncertainty calibration, raising concerns in high-stakes scenarios~\cite{leung2022}.
Moreover, their black-box nature hinders transparency, auditability, and alignment with established clinical guidelines~\cite{holzinger2019, diaz2022}.

\paragraph{Trustworthy AI Systems}
To address the LLM's limitations, several methods have been proposed, such as prompt engineering~\cite{liu2023}, retrieval-augmented generation (RAG)~\cite{sun2023}, and reinforcement learning from human feedback (RLHF)~\cite{ouyang2022}.
Hybrid approaches that combine LLMs with structured medical ontologies or external knowledge bases (e.g., UMLS, SNOMED-CT) have also been explored~\cite{le2025}.
Recent work additionally explores multimodal integration, such as combining imaging and text for radiology reports~\cite{chambon2022}, and aligning LLM reasoning with clinical logic using symbolic constraints~\cite{lucas2024}.
Despite these efforts, current systems still fall short of fully trustworthy, interpretable, and verifiable clinical decision support. 

\paragraph{Neuro-Symbolic Systems}
Recent advances have extended symbolic reasoning with fuzzy logic and probabilistic rules to better model uncertainty in clinical settings~\cite{zadeh1996, lukasiewicz2008}. 
However, these systems suffer from limited scalability and brittleness when faced with the variability and ambiguity inherent in natural language narratives.
To bridge the gap between data-driven learning and logical reasoning, neuro-symbolic systems have gained traction in recent years. 
Approaches such as Neural Theorem Provers~\cite{rocktaschel2017}, Logic Tensor Networks~\cite{serafini2016}, and DeepProbLog~\cite{manhaeve2018} enable end-to-end differentiable reasoning over symbolic structures. 
In the medical domain, some work has explored integrating neural embeddings with ontology-driven systems to improve interpretability~\cite{choi2017}. 
Nonetheless, most existing neuro-symbolic methods either lack support for fuzzy and probabilistic reasoning or do not handle real-time physician-in-the-loop feedback.

Unlike prior neuro-symbolic systems that aim to integrate neural perception
with symbolic world models,
our work leverages formal methods as a normative and verification layer.
We do not assume complete or closed-world symbolic knowledge,
nor do we rely on symbolic inference as the primary decision mechanism.
\section{Conclusion}

We presented a neuro-symbolic reasoning framework that integrates LLMs, logical inference, fuzzy quantification, and probabilistic modeling for interpretable and robust clinical diagnosis.
By combining symbolic rules with data-driven mechanisms, our system bridges the gap between discrete knowledge representation and real-world medical data's uncertain, continuous nature.
Our approach enables transparent diagnostic inference under partial or vague observations, offering human-readable reasoning chains and calibrated confidence scores.
Through experiments on public benchmarks, we demonstrate that the integration of LLMs and logic improves explainability, flexibility, and robustness. Our work contributes to the growing effort to build AI systems that are not only accurate but also interpretable, robust, and aligned with human reasoning---an essential step toward safe deployment in complex real-world environments.

\newpage

\bibliographystyle{plain}
\bibliography{references}

\appendix

\newpage

\end{document}